\documentclass[letterpaper]{article}

\usepackage{aaai}
\usepackage{times}
\usepackage{helvet}
\usepackage{courier}

\begin{document}

\title{Dialogue as Discourse: Controlling Global Properties of Scripted Dialogue\thanks{We would like to thank David Arnold, Richard Power and two anonymous reviewers for helpful comments. This research is supported by the EC Project NECA IST-2000-28580. The information in this document is provided as is and no guarantee or warranty is given that the information is fit for any particular purpose. 
The user thereof uses the information at their own risk and liability.}}
\author{Paul Piwek \& Kees van Deemter\\
ITRI - University of Brighton\\
Watts Building, Moulsecoomb\\
BN2 4GJ Brighton, UK\\
\{Paul.Piwek, Kees.van.Deemter\}@itri.bton.ac.uk}

\maketitle

\begin{abstract}
\begin{quote}
This paper explains why {\em scripted dialogue} shares some 
crucial properties with discourse. In particular, when 
scripted dialogues are generated by a Natural Language 
Generation system, the generator can apply revision 
strategies that cannot normally be used when the dialogue 
results from an interaction between autonomous agents
(i.e., when the dialogue is not scripted). The paper 
explains that the relevant revision operators are best 
applied at the level of a dialogue plan and discusses 
how the generator may decide when to apply a given 
revision operator.
\end{quote}
\end{abstract}

\section{Controlling Global Properties in Text Generation}

A Natural Language Generation ({\sc nlg}) system puts information into words.
When doing this, the system makes a large number of {\em decisions} as to
the specific way in which this is done: aggregating information into
paragraphs and sentences, choosing one syntactic pattern instead of
another, deciding to use one word rather than another, and so on.
For many purposes, such decisions can be made on the basis of local
information. The choice between an active and a passive sentence, for
example, usually does not take other decisions (such as the analogous
choice involving another sentence) into account. In slightly more
difficult cases, the decisions of the generator can be based on
decisions that have been taken {\em earlier}. For example,
the generator may inspect the linguistic context to the left of the
generated item for deciding between using a proper name,
a definite description, or a personal pronoun.  There are, however,
situations in which generative decisions require information about text
spans that have not yet been generated. Such situations typically arise
when the generated text is subject to global constraints, i.e.,
constraints on the text as whole, such as its length. For example,
suppose there is length constraint. Now, in order to decide
whether it is necessary to use aggregation when generating some subspan
(to stay beneath the maximum length), it is necessary to know the length of
the text outside of the current subspan. This can lead to
complications since at the point in time when the current span is
generated, the rest of the text might not yet be available.\footnote{
Compare the constraint, in daily life, that a party has to {\em fit into 
one's living room}. This constraint can lead to interdependencies 
between decisions. For instance, suppose the room can contain
ten visitors, and eleven invitations have already gone out. Is it 
wise to send out another invitation? This depends on how many of the
invitations will be accepted, which can be difficult to foresee. 
Another, more qualitative contraint may be that the party has to 
{\em be pleasant}, and this may introduce further interdependencies 
(e.g., if Mr. $x$ comes to the party then Mrs. $y$ had better not 
be invited).}
One class of global {\sc nlg} constraints involves the linguistic style of
a text, as regards its degree of formality, detail, partiality, and so
on (Hovy 1988). Whether a text can be regarded as {\em moderately
formal}, for instance, depends on whether formal words and patterns
are chosen {\em moderately often}, which makes this a quintessentially
global constraint. Moreover, style constraints may contradict each
other. For example, if a text has to be {\em impartial} as well as {\em
informal} then the first of these constraints may be accommodated by 
choosing a passive construction, but the second would be tend to be
adversely affected 
by that same choice. Hovy argues that these problems make top-down 
planning difficult because it is hard to foresee what `fortuitous 
opportunities' will arise during later stages of {\sc nlg}. Perhaps more
contentiously, he argues that these problems necessitate a {\em
monitoring} approach, which keeps constant track of the degree to 
which a given text
span satisfies each of a number of constraints (e.g., low
formality, low partiality).  After generating the first $n$ sentences
of the text, the remainder of the text is generated in a way that takes
the degrees of satisfaction for all the style constraints into account,
for example by favouring those constraints that have been
least-recently satisfied (or least-often satisfied of the text span
in its totality). Monitoring is an attempt to address
global constraints in an incremental fashion and may be
viewed as a plausible model of spontaneous speech. It may be
likened to steering a ship: when the ship is going off course,
you adjust its direction. Needless to say, there is no guarantee 
that monitoring will result in a happy outcome, but it is a
computationally affordable approach to a difficult problem. 

As we have seen, another example of a global constraint is the constraint
to generate a text of a specified {\em size} (e.g., in terms of the
number of words or characters used). Reiter (2000) discusses various
ways in which the size of a generated text may be controlled. We
already saw that for certain generation decisions it might be necessary
to know the length of the remainder of the text. Based on experiments
with the {\sc stop} system, Reiter observes that the size of a text is
difficult to {\it estimate} (i.e., predict) on the basis of an abstract
document plan. He argues that {\em revision} of texts, based on
measuring the size of an actually generated text, is the way to go:
when the size measure indicates that the text is too large, the least
important message in the sentence plan was deleted and the text
regenerated.

Revision has been claimed to be a good model of human {\em writing}
(e.g., Hayes and Flower 1986).  In some respects, size is a particularly
difficult kind of global constraint, since it applies to the form
rather than the content of a text, and this is why Reiter and
colleagues let revision wait until a draft of the text has been
generated and evaluated (i.e., after its size has been measured).
%
%
In other respects, it is relatively simple, because it is
one-dimensional and straightforward to define.

In this paper, we discuss a class of non-local decisions in the generation
of dialogue that were first introduced in Piwek and van Deemter (2002). To be able to make these decisions in a well-motivated
way, without complicating the {\sc nlg} system unduly, we adopt a revision
approach, which makes our system more similar to Reiter's {\sc stop}
system than to Hovy's {\sc pauline}. Like {\sc stop}, we  use a
revision strategy, but unlike {\sc stop}, we do not need to evaluate
the textual output of the generator since, in our case, evaluation can
be done on the basis of an abstract representation of the dialogue
content.

\section{Non-local decision making in Dialogue Generation}

Let us define more precisely what we mean by non-local decisions
in generation:
\begin{quote}
({\sc non-local decisions in generation}) Given a generator $G$ that is
producing some subspan $s_a$ of a document $D$, we will call a decision
by $G$ concerning the generation of $s_a$ {\em non-local} if it
requires information regarding the content and/or realization  of some
other span $s_b$ in $D$, where some or all of $s_b$ is not included in
$s_a$.
\end{quote}
\noindent
Given a left-to-right generation algorithm, there is no problem if
$s_b$ precedes $s_a$: content and form of $s_b$ can be stored so that
they are available when $s_a$ is generated. However, if all or part of
$s_b$ succeeds $s_a$, (i.e., has not yet been generated), we have a
problem. In the preceding section, three solutions for this problem have
been exemplified: one was based on {\em estimating} the relevant
properties of $s_b$, one on {\em revising} $s_a$, and one on 
constantly monitoring the satisfaction of the global properties 
that need to be satisfied. 

Note that revision is a strategy which has to be treated with 
care. For instance, we need to make sure that never-ending 
cycles of revision do not occur. Such a cycle would, for 
instance, occur if a particular revision operation always 
created circumstances which allowed it to be applied again. We 
return to this issue later on in this paper. 

In the remainder of this paper we discuss what options are available if the
text that needs to be generated is a dialogue. 

Investigations into dialogue by
computational linguists have typically focussed on the communication
between the interlocutors that take part in the dialogue. The dialogue
is taken to consist of communicative interaction between two
individuals. There are, however, many situations where a
dialogue fulfils a different purpose: dialogues in the theatre, on
television (in drama, sitcoms, commercials, etc.) and on the radio (in
radio plays, commercials, etc.) are not primarily meant as vehicles
for genuine communication between the interlocutors, but rather at aiming to
{\it affect the audience} of the dialogue (in the theatre, in front of
the {\sc tv} or radio). The effects are often achieved through global
properties of the dialogue (the dialogue should take only a certain 
amount of time, should be entertaining, should teach the 
audience something, should make a certain point forcefully, 
etc.). In short, if we look at the dialogue from the perspective of 
an audience, global properties of the dialogue are of great importance.

\subsection{The Information State approach}

Firstly, let us consider the currently prevalent approach to generating
dialogue.  The starting point is the use of two autonomous agents, say
$A$ and $B$. These agents are associated with Information States (e.g.,
Traum et al., 1999): $IS(A)$ and $IS(B)$. The agent who holds the turn
generates an utterance based on {\it its} Information State. This leads
to an update of the Information States of both agents. Subsequently,
whichever agent holds the turn in the new Information States, produces
the next utterance. In most implemented systems, one of the agents is a
dialogue system and the other a human user; the approach has, however,
also been used for dialogue simulations involving two or more software
agents (as pioneered by Power, 1979).

For our purposes, it is important to note that the agents only have
access to their own Information State, and can only use this state to
produce contributions. This has some repercussions when it comes to
controlling global properties of dialogue. Estimation becomes more
difficult if it involves a span that is to be generated by the other
agent. The agent has no access to the Information State of this agent
and will therefore find it more difficult to estimate what it is going
to say.  Furthermore, estimating one's own future utterances can become
more difficult, since they may succeed and depend on utterances by the
other agents. In general estimation for the purpose of
coordinating the generation of the current span with not yet generated
spans is more difficult/less reliable in the Information State approach.

Revision is also much more limited if the Information State approach is
strictly followed. Revision is only possible within a turn. The
Information State approach assumes that turns are produced according to
their chronological ordering, and hence it is not possible to go back
to a turn once it has been completed.
%

\subsection{The Dialogue Scripting approach}

An alternative to the Information State approach is the dialogue
scripting approach. In Piwek \& Van Deemter (2002) we take the main
characteristic of this approach to be that it involves the creation of
a dialogue (script) by one single agent, i.e., the script author. Thus,
the production of the dialogue is seen as analogous to single-author
text generation. 

The automated generation of scripted dialogue has been poineered by 
Andr\'e et al. (2000). We follow Andr\'e et al. (2000) in distinguishing
between the creation of the dialogue text, i.e., the script and the
performance of this script. Of course the Information State approach
also lends itself for such a separation, but typically the authoring
and performance functions are taken care of by the same agent (the
interlocutors) and take place at the same time. In the scripting
approach, the script for the entire dialogue is produced first. The
performance of this script takes place at a later time, typically by
actors who are different from the author.

There are at least two reasons why the scripting approach is better
suited to creating dialogues with certain global properties than
Information State approaches. Firstly, in the scripted dialogue
approach the information and control resides with one author. This
makes estimation more reliable; assuming that it is easier to predict
one's own actions than those of another agent. Secondly, the scripting
approach does not presuppose that the dialogue is created in the same
temporal order in which it is to be performed. Hence it is possible to
revisit spans of the dialogue and edit them.

In between traditional Information State and Scripted Dialogue
approaches, {\it hybrid} approaches are possible. For instance, one might
generate a dialogue according to the Information State approach, and
then edit this draft with a single author.
The techniques described in the next section are presented in the
context of pure Dialogue Scripting but they remain largely 
valid for more hybrid set-ups.

Despite the existence of hybrid approaches it is important to keep in
mind the different perspectives from which Information State and
Scripted Dialogue approaches arose: the Information State approach
focuses on the communication between the interlocutors in
the dialogue, whereas the scripted dialogue approach focuses on the
communication between the script author and the readers/audience of the
dialogue; the communication between the interlocutors of the scripted
dialogue is only pretended communication.

\section{Exploring the Control of Global Dialogue Properties in NECA}

In this section we discuss the control of global
dialogue properties in the {\sc neca} system.\footnote{{\sc neca} is an
EU-IST funded project which started in October 2001 and has a duration
of 2.5 years. {\sc neca} stands for `Net Enviroment for Embodied
Emotional Conversational Agents'. The following partners are involved
in the project: {\sc Dfki}, {\sc Ipus} (University of the Saarland),
{\sc Itri} (University of Brighton), {\sc \"ofai} (University
of Vienna), Freeserve and Sysis {\sc ag}. Further details can be 
found at: http://www.ai.univie.ac.at/NECA/.} {\sc neca} generates
Dialogue Scripts that can be performed by animated characters.
Currently, a prototype exists --called eShowroom-- for the generation
of car sales dialogues; a prototype for a second domain, Socialite,
involving social chatting is under construction. Here we focus
on eShowroom, which will be featured on an internet portal for car
sales. 

\subsection{The NECA eShowroom system} 

The eShowroom demonstrator allows a user to browse a database of 
cars, select a car,
select two characters and their attributes, and subsequently view an
automatically generated film of a dialogue between the characters about
the selected car. The eShowroom system is provided with the following
information as its input:
%
\begin{itemize}
\item A database with facts about the selected car (maximum speed, horse
power, fuel consumption, etc.).
\item A database which correlates facts with value dimensions such as 
`sportiness', `environmental-friendliness', etc. (e.g., a high maximum 
speed is good for `sportiness', high gasoline consumption is bad for 
the environment).
\item Information about the characters:
  \begin{itemize}
  \item Personality traits such as extroversion and agreeableness. 
  \item Personal preferences concerning cars (e.g., a preference 
        for cars that are friendly for the environment).
  \item Role of the character (either seller or customer).
  \end{itemize}
\end{itemize}
\noindent
This input is processed in a pipeline that
consists of the following modules:
\begin{enumerate}
\item A Dialogue Planner, which produces an abstract 
description of the dialogue (the dialogue plan).
\item A multi-modal Natural Language Generator which specifies 
linguistic and non-linguistic realizations for the dialogue acts 
in the dialogue plan.
\item A Speech Synthesis Module, which adds information for Speech.
\item A Gesture Assignment Module, which controls the temporal 
coordination of gestures and speech.
\item A player, which plays the animated characters and the corresponding
speech sound files. 
%
\end{enumerate}
\noindent
Each step in the pipeline adds more concrete information to the
dialogue plan/script until finally a player can render it (see also 
Krenn et al., 2002). A single
{\sc xml} compliant representation language, called {\sc rrl}, has been
developed for representing the Dialogue Script at its various stages of
completion (Piwek et al., 2002).

The following is a transcript of a dialogue fragment which the system 
currently generates (Note that this is only the text. The system 
actually produces spoken dialogue accompanied by gestures of the 
embodied agents which perform the script):

\begin{quote}
{\sc Seller:} Hello! How can I help? \\
{\sc Buyer:} Can you tell me something about this car? \\
{\sc Seller:} It is very comfortable. \\
{\sc Seller:} It has leather seats. \\
{\sc Buyer:} How much does it consume? \\
{\sc Seller:} It consumes 8 liters per 60 miles. \\
{\sc Buyer:} I see. \\
{\sc Etc.}
\end{quote}

\noindent
Here, we focus on the representation of this dialogue after 
it has been processed by the Dialogue Planning module. The 
{\sc rrl} dialogue script consists of four parts: 

\begin{quote}
1. A representation of the initial common ground of the 
interlocutors. This representation provides information 
for the generation of referring expressions.  
\end{quote}

\begin{quote}
2. A representation of each of the participants of the dialogue. 
For instance, we have the following representation for the seller 
named Ritchie: 

\begin{verbatim}
  <person id="ritchie">
      <realname firstname="Ritchie"
       title="Mr"/>
      <gender type="male"/>
      <appearance character=
        "http://neca.sysis.at/eroom/
         msagent/ritchie_hq/
         Ritchie_Off.acf"/>
      <voice name="us2">
        <prosody pitch="-20%" 
         rate="-10%"/>
      </voice>
      <personality 
         agreeableness="0.8" 
         extraversion="0.8"
         neuroticism="0.2" 
         politeness="polite"/>
      <domainSpecificAttr 
         role="seller" 
         x-position="70" 
         y-position="200"/>
  </person>
\end{verbatim}
\end{quote}

\begin{quote}
3. A representation of the dialogue acts. Each act is associated 
with attributes, some of which are optional, specifying its type, 
speaker, addressees, semantic content (in terms of a discourse 
representation structures, Kamp \& Reyle, 1993), what it is a 
reaction to (in terms of conversation analytical adjacency pairs) 
and the emotions with which it is to be expressed. The following 
is the representation for the dialogue act corresponding
with the sentence `It has leather seats':

\begin{verbatim}   
  <dialogueAct id="v_4">
      <domainSpecificAttr 
            type="inform"/>
      <speaker id="ritchie"/>
      <addressee id="tina"/>
      <semanticContent id="d_4">
        <drs id="d_3">
          <ternaryCond 
            argOne="x_1" 
            argThree="true"
            argTwo="leather_seats" 
            id="c_4" 
            pred="attribute"/>
        </drs>
      </semanticContent>
      <reactionTo id="v_3"/>
  </dialogueAct>
\end{verbatim}
\end{quote}

\begin{quote}
4. The fourth component of the {\sc rrl} representation of the 
dialogue script consist of the temporal ordering of the dialogue 
acts:

\begin{verbatim}
  <temporalOrdering>
      <sequence>
        <act id="v_1"/>
        <act id="v_2"/>
        <act id="v_3"/>
        <act id="v_4"/>
             .
             .
             .
      </sequence>
  </temporalOrdering>
\end{verbatim}

\end{quote}

\subsection{Enforcing global constraints in dialogue}
 
Here we want to examine how to adapt the system so that it can take
global constraints into account. We have seen in the preceding section 
that the {\em Dialogue Scripting} approach is best suited
for the control of global dialogue properties. The {\sc neca} system is
based on Dialogue Scripting: each module in the pipeline
operates as a single author/editor which creates/elaborates the
Dialogue Script.
Within the Dialogue Scripting approach various methods for controlling
global properties can be employed. Earlier on, we discussed monitoring,
estimation and revision as approaches that have been employed in text
generation. For the purpose of this paper, we limit our
attention to the revision approach.

There are a number of reasons for this choice. Firstly, monitoring
is tailored to left-to-right processing, whereas the dialogue 
scripting approach is not constrained in this way. Moreover, 
monitoring can only work well if the number of decisions relating 
to each constraint is very large, since this gives the system
many opportunities for `changing course'. But even if a single-pass 
monitoring approach {\em could} work well at the level of dialogue, it 
would tend to complicate the design and maintainance of the system 
(cf. Callaway \& Lester 1997, Reiter 2000, Robin \& 
McKeown 1996).\footnote{We could, for instance, have
included the aggregation and insertion operations (see below)
directly in our dialogue manager, but this would have complicated 
the dialogue planner rules.} Having a
separate revision module also allows for a more straightforward
division of labour between multiple system developers. (In
{\sc neca}, for example, the dialogue planner and the revision 
system are developed at different sites.) 

Furthermore, Reiter (2000) argues that revision compares 
favourably with other techniques for satisfying size constraints 
(a specific type of global constraint) in text generation. He 
compares revision with heuristic size estimators (for predicting 
the size of the text on the basis of the message to be conveyed) 
and multiple solution pipelines. Robin and McKeown (1996) discuss 
the implementation of revision in a summarization system for 
quantitative data ({\sc streak}). They carried out two evaluations 
which they argue show 
that their revision-based approach covers a significantly larger
amount of the structures found in their corpus than a traditional
single-pass generation model. Additionally, they claim that their 
evaluation shows that it is easier to extend the revision-based 
approach to new data.

Our approach to revision differs from other 
approaches. Firstly, our revisions are carried out on the
abstract dialogue plan, before linguistic realization. Although
Callaway \& Lester also carry out their revision operations on abstract
representations of sentences, these are obtained by first generating
concrete sentences and then abstracting again over irrelevant details.
Instead of first fully generating and then abstracting, we follow an
approach of partial generation. Secondly, Reiter and
Callaway \& Lester focus on a single type of constraint. In
this respect, our work is more
similar to that of Hovy, where different types of potentially
conflicting constraints are considered. To our knowledge, we
are the first to propose revision operations on {\it dialogue}
structure as opposed to discourse or sentence structure. Ultimately,
of course, these different types of revision ought to be addressed 
through one common approach.

\subsection{Two constraints and a revision problem}

To illustrate the issues, let us consider two global
constraints on dialogue:
\begin{itemize}
\item Number of turns in the dialog ({\sc Turns}): maximal ({\sc
max}) or minimal ({\sc min})
\item Degree of Emphasis ({\sc Emph}): maximal ({\sc max}) or min
({\sc min})
\end{itemize}  
\noindent
For the moment, we keep the constraints as simple as possible 
and assume that they can only take extreme values ({\sc max} or {\sc min}).  

Furthermore, we introduce two revision operations on the output 
of the dialogue planner:
\begin{itemize}
\item {\sc Adjacency Pair Aggregation (Aggr)}
\begin{quote}
{\bf Operation}: Given the adjacency pairs $A$ = ($A_1$,$A_2$) and $B$
= ($B_1$,$B_2$) in the input, create A+B = ($A_1$+$B_1$,$A_2$+$B_2$).\\[2ex] 
{\bf Precondition}: $A$ and $B$ are about the same value dimension.\\[2ex] 
{\bf Example}: $A$ = (Does it have airbags? Yes), \\
B = (Does it have
ABS? Yes), \\ A+B = (Does it have airbags and ABS? Yes) \\[2ex] {\bf Comment}:
The shared value dimension is {\em security}.
\end{quote}
\item {\sc Adjacency Pair Insertion (Insert)} 
\begin{quote}  
{\bf Operation}: Given adjacency pair $A$ = ($A_1$,$A_2$) in the input,
1. create adjacency pair $B$ = ($B_1$,$B_2$) which is a clarificatory
subdialogue about the information exchanged in $A$ and 
2. insert $B$ after $A$, resulting in ($AB$) = ($A_1$,$A_2$)($B_1$,$B_2$).\\[2ex] 
{\bf Precondition}: The information exchanged in A is marked for emphasis.\\[2ex] 
{\bf Example}: $A$ = (Does it have leather seats? Yes). Assume that
{\em comfort} is positively correlated with having leather seats and that the
user has indicated that the customer prefers comfortable cars. On the
basis of this, the information exchanged in A is marked for emphasis.
The text after revision is: ($AB$) = Does it have leather seats? Yes.
Real leather? Yes, genuine leather seats.\\[2ex]
{\bf Comment}: Piwek \& Van Deemter (2002) contains examples of how 
human authors of scripted dialogue appear to use sub-dialogues for 
emphasis.

\end{quote}
\end{itemize}

\noindent
In the definitions of the two operations we use the notion of an 
{\em adjacency pair} which is common in Conversation Analysis. 
The idea is that the
first and second part of the pair are connected by the relation of
conditional relevance (e.g., a pair consisting of a question and an
answer): `When one utterance (A) is conditionally relevant to another
(S), then the occurrence of S provides for the relevance of the
occurrence of A' (Schegloff, 1972:76).

Let us now describe our revision problem. We have an initial dialogue
plan $dp_1$, produced by the dialogue planner. Before it is passed on to
the multi-modal natural language generator we want to apply the
revision operations {\sc Aggr} and {\sc Insert} in such a way that the
resulting dialogue plan $dp_2$ optimally satisfies the constraints for
{\sc turn} and {\sc emph}. In total, there are four possible constraint
settings: 

\begin{enumerate}
\item {\sc turn = max} and {\sc emph = max}.
\item {\sc turn = max} and {\sc emph = min}.
\item {\sc turn = min} and {\sc emph = min}.
\item {\sc turn = min} and {\sc emph = max}.
\end{enumerate}

\subsection{Sequential revision}

Let us look at two alternative ways in which these constraint settings
might be satisfied. The first is simple-minded but efficient: First,
one operation is applied as often as needed and then the same is done
for the other operation. We assume that insertion correlates with {\sc
emph} and aggregation correlates with {\sc turn}.  If a constraint is
set to {\sc max}, the operation is performed as often as possible; if
the constraint is set to {\sc min}, the operation is not applied at 
all. Note that this procedure will always terminate, given that our initial
dialogue plan contains only a finite number of pieces of information
that are marked for emphasis and there is only a finite number of
adjacency pairs which share a value dimension. Hence the preconditions
of the operations can only be satisfied a finite number of
times.

There are, however, complications, since constraints may be
interdependent. One type of problem obtains when one
operation affects (i.e. creates or destroys) the preconditions 
for another. Suppose, for example, 
our setting is ({\sc turn = max}, {\sc emph = max}), while aggregation 
is performed before insertion. In this case, a less than maximal number 
of aggregations would result, since insertion can introduce new candidates 
for aggregation. This type of problem can usually be finessed by
finding an `optimal' ordering between operations: if insertion 
preceeds aggregation, both constraints of our example situation
can be satisfied. Unfortunately, however,
there is another type of problem which cannot be finessed so easily.
Suppose, for example, our setting is ({\sc turn = min}, {\sc emph = max}),
and observe that insertion positively affects the number of turns
as well as the degree of emphasis, affecting the two constraints 
({\sc turn = min}, {\sc emph = max}) in
opposite ways. In such a case it is unclear what the best strategy is: 
the algorithm might either maximize the number of insertions, trying 
to maximize emphasis, or minimize them, trying to
minimize the number of turns. To tackle problems of both kinds,
an approach is needed that is able to make {\em trade-offs} between 
conflicting constraints.

\subsection{A `Generate and Test' approach} To tackle both these problems,
we propose a `generate and test' approach to the revision problem. The 
algorithm proceeds as follows:

\begin{enumerate}
\item We use a conventional topdown planner to produce a single 
dialogue plan $dp_{start}$. 
\item Next, we generate all
possible plans that can be obtained by applying the operations
{\sc Insert} and {\sc Aggr} zero or more times, in any order, to 
$dp_{start}$. Let us
call this {\it set} of all possible output plans $DP_{out}$.
\item Each member of
$DP_{out}$ is assigned a score for the {\sc turn} and {\sc emph} constraints.
Each dialogue plan $dp \in DP_{out}$ is characterized as a tuple
consisting of the {\sc turn} and {\sc emph} scores $\langle S_T, S_E
\rangle$. The {\sc turn} score $S_T$ depends on the number of turns of
the dialogue plan. The {\sc emph} score $S_E$ depends on the number of
emphasis subdialogues in the dialogue plan that were added during 
revision. We assume that our scores
are normalized, so they each occupy a value on the interval [0-100],
i.e., satisfaction of the constraint from $0\%$ to $100\%$. $100\%$
means that there is no alternative $dp$ which does better.
\item Finally, on the basis of the scores assigned to the plans 
in $DP_{out}$ and according to some arbitration plan, we select 
an optimal outcome or set of optimal outcomes, i.e., a unique 
solution or a set of solutions.
\end{enumerate}

\noindent
At this point, one might ask how we decide which operations are 
included in the conventional topdown planner and which ones are 
deemed revision operations. To answer this question, it will be 
useful to elaborate a bit on our underlying assumptions about 
dialogue. 

In classical Discourse Theory (e.g.,. Stenstr\"om, 1994) 
conversations typically consist of three 
distinguishable phases: an opening, a body and a closing.
each of which has a specific purpose.
According to for, instance, Clark (1996) individual dialogue 
acts belong to different tracks depending on how directly they 
contribute to the purpose of the dialogue phase in which they 
occur. On track $1$ we have dialogue acts which are intended 
to immediately contribute to furthering this purpose, for instance, 
the buyer's asking for the price of the car to the seller or 
the buyer's introducing him- or herself to the 
seller.\footnote{
In our tentative view, greetings occur at the level of 
track $1$, since they directly 
further the purpose of the opening and are, therefore,
distinct from the metacommunication which takes place 
on track $2$. Alternative views would be equally easy to 
model.} Metacommunication about the communication 
on track $1$ takes place at the level track $2$. This 
includes monitoring the success of the communication, 
attempting to fix communication problems, etc. 

For our purposes, the distinction between acts on track 
$1$ and $2$ is a useful one, since acts on track $2$ can be 
viewed as mere decorations of the acts which further the 
purpose of the conversation on track $1$. For example, if we 
omit the utterances on track $2$ the remaining dialogue script 
still makes sense (cf. Piwek \& Van Deemter, 2002), whereas 
removing utterances from track $1$ does not have the same 
effect. Consider, for instance, the following exchange:

\begin{quote}
1. {\sc Buyer}: How much does the car cost? \\
2. {\sc Seller}: 15.000 Euro. \\
3. {\sc Buyer}: 15.000? \\
4. {\sc Seller}: Yes, only 15.000. 
\end{quote}

\noindent
The acts on track $1$ (1. and 2.) make sense on their own, whereas 
those on track $2$ (3. and 4.) do not. For this reason, acts on 
track $1$ are dealt with by the dialogue planner, while acts on 
track $2$ are inserted at the revision stage, by means of the 
operation {\sc Insert}.

The operation {\sc Aggr} is an instance of an aggregation operation 
on the dialogue level. Aggregation operations are typically dealt 
with as involving revision: two or more structures are merged/revised 
into one new structure. Our {\sc Aggr} allows us to reorganize the 
location of dialogue acts. It does not add or remove any dialogue 
acts. The precondition on {\sc aggr}, which stipulates that only 
dialogue acts which deal with the same value dimension can be 
aggregated, guards against erratic reorganizations of the dialogue, 
destroying smooth shifts from one topic (value dimension) to 
another.\footnote{More generally, the generation of texts with 
smooth topic shifts can be seen as a constraint satisfaction 
problem. See, for instance, Kibble \& Power (2000).}
 
A second issue which the current sketch for an algorithm 
raises is that of the choice of an `arbitration plan' for 
selecting the solution or set of solutions from $DP_{out}$. 
Fortunately, this is a well-known problem in decision theory 
and more specifically game theory. One would like an arbitration 
plan to satisfy certain criteria which define a fair balancing 
of different constraints. One set of such criteria is due 
to John Nash, who proposed that a solution should 
satisfy the following four axioms (Nash, 1950):

\begin{enumerate}
\item {\sc Linear Invariance}: If one transforms the scores for 
either constraint by a positive linear function, then the 
solution should be subject to the same transformation. This 
axiom derives from the fact that the score/utility functions 
in game theory are normally taken to be an interval scale. 
These are invariant only under positive linear transformations. 
\item {\sc Symmetry}: If for each outcome associated with a 
pair of scores $\langle x, y \rangle$, there is another 
outcome $\langle y, x \rangle$, then the solution should 
consist of a pair of identical scores $\langle z, z \rangle$. 
In words, constraints are treated as equals with respect to 
each other.
\item {\sc Independence of Irrelevant Alternatives}: 
Suppose we have two different outcome sets $A$ and $B$. 
Assume also that $A \subset B$. If the solution for $B$ 
is a member of $A$, then this solution should also be a 
solution for $A$. In words, the unavailability of non-solution 
outcomes should not influence the final solution.  
\item {\sc Pareto Optimatily}: The solution should be Pareto 
Optimal.  A pair $\langle S_T,S_E \rangle \in DP_{out}$
is Pareto Optimal iff it is impossible to find another pair
$\langle S_T',S_E' \rangle \in DP_{out}$ such that:
\begin{enumerate}
\item $S_T'$ = $S_T$ and $S_E'$ $>$ $S_E$ or 
\item $S_E'$ = $S_E$ and $S_T'$ $>$ $S_T$ or
\item $S_E'$ $>$ $S_E$ and $S_T'$ $>$ $S_T$.
\end{enumerate}
\end{enumerate}
\noindent
For our purpose, the notion of Pareto Optimality is 
particularly interesting. A pair is Pareto Optimal if and 
only if it is impossible to improve
one of its elements without making the other element worse 
off. Unfortunately, Pareto Optimality does not help us to 
identify a unique solution: For example, if $DP_{out}$ 
contains only two
pairs: $dp_1 = \langle 100,10 \rangle$ and 
$ dp_2 = \langle 50,50 \rangle$ then both pairs are
Pareto Optimal. 

Nash came up with an arbitration plan which satisfies not 
only Pareto  Optimality, but all four of the proposed axioms. 
The idea is that by satisfying the four axioms, Nash's 
arbitration plan provides a `fair' solution to the problem 
of maximizing the degree to which both 
(in general: all) contraints are satisfied, that is, a 
solution that treats the two constraints evenhandedly. 
According to the Nash 
arbitration plan, the optimal solution is the solution 
$\langle S_T,S_E \rangle$ with the highest value for 
$S_T \times S_E$. This plan causes $dp_2$ to 
win, which is a desirable outcome, in our 
opinion.

The Nash plan is 
guaranteed to choose a solution that is Pareto Optimal.
The same is true for a plan that maximizes the sum 
instead of the product of the scores, but this would fail to
punish a treatment that favoured one constraint over the
other, as in the case of the $dp_1 = \langle 100,10 \rangle$ and 
$ dp_2 = \langle 50,50 \rangle$. 

\section{Conclusions}

We have discussed various existing methods for 
controlling global properties of generated text (such as length
and style). Having done this, we have focused on generated
{\it dialogue}, offering a number of arguments 
in favour of the {\em Scripted} Dialogue approach. Building on 
observations in the literature, we went on to make a case for 
{\em revision} approaches to controlling global properties of 
scripted dialogue. More specifically, we have sketched
the potential for a revision approach to dialogue generation
in the {\sc neca} system. In our discussion of {\sc neca}, we have 
outlined a new `generate and test' approach that would put
well-known techniques from decision theory and game theory to a 
novel use, thereby exemplifying a recent trend in 
formal and computational linguistics (Rubinstein 2000).

\section{References}

\small{
\begin{description}

\item[] Andr\'{e}, E., T. Rist, S. van Mulken, M. Klesen \& S. Baldes (2000). `The Automated Design of Believable Dialogues for Animated Presentation Teams'. In: J. Cassell, J. Sullivan, S. Prevost and E. Churchill, {\em Embodied Conversational Agents}, MIT Press, 220-255. 

\item[] Callaway, C. and J. Lester (1997). `Dynamically Improving Explanations: a Revision-Based Approach to
Explanation Generation'. In: {\it Proceedings. of IJCAI97 conference}, Nagoya, Japan.

\item[] Clark, H. (1996). {\em Using Language}. Cambridge University Press, Cambridge.

\item[] Hayes, J. and L. Flower (1986). `Writing research and the writer'. {\em American Psychologist} {\bf 41}, 1106--1113.

\item[] Hovy, E. (1988). {\em Generating Natural Language Under Pragmatic Constraints}. Lawrence Erlbaum Associates, Hillsdale, New Jersey.

\item[] Kamp, H. \& Reyle, U. (1993). {\it From Discourse
to Logic: Introduction to Modeltheoretic Semantics of Natural
Language, Formal Logic and Discourse Representation Theory}. Kluwer Academic Publishers, Dordrecht.

\item[] Kibble, R. \& R. Power (2000). `An integrated framework for text planning and pronominalisation'. In: {\it Proceedings of The First International Natural Language Generation Conference (INLG'2000)}, 77--84

\item[] Krenn B., H. Pirker, M. Grice, S. Baumann, P. Piwek, K. van Deemter, M. Schr\"oder, M. Klesen, E. Gstrein (2002). `Generation of multimodal dialogue for net environments', in: Busemann S. (ed.), {\it KONVENS 2002}, DFKI, Saarbr\"ucken, Germany, 91--98.

\item[] Nash, J. (1950). `The Bargaining Problem'. {\it Econometrica}, {\bf 18}, 155--162.

\item[] Piwek, P. and K. van Deemter (2002). `Towards Automated
Generation of Scripted Dialogue: Some Time-Honoured Strategies'. In:
Bos, J., M. Foster and C. Matheson, {\it Proceedings of EDILOG: 6th
workshop on the semantics and pragmatics of dialogue},
Edinburgh, September 4-6, 2002.

\item[] Piwek, P., B. Krenn, M. Schr\"oder, M. Grice, S. Baumann and H. Pirker (2002). `RRL: A Rich Representation Language for the Description of Agent Behaviour in NECA'. In: {\it Proceedings of the AAMAS workshop "Embodied conversational agents - let's specify and evaluate them!"}, Bologna, Italy, 16 July 2002.

\item[] Power, R. (1979). `The Organization of Purposeful Dialogues'.
{\em Linguistics}, {\bf 17}, 107--152.

\item[] Reiter (2000). `Pipelines and Size Constraints'. 
{\it Computational Linguistics}. {\bf 26}:251-259.

\item[] Robin, J. \& K. McKeown (1996). `Empirically Designing and Evaluating a New Revision-Based Model for Summary Generation', {\it Artificial Intelligence}. {\bf 85}(1-2).

\item[] Rubinstein, A. (2000). {\em Economics and Language}, Cambridge University Press, Cambridge.

\item[] Schegloff, E. (1972). `Notes on a Conversational Practice:
Formulating Place'. In: D. Sudnow (ed.), {\it Studies in Social
Interaction}, The Free Press, New York, 75--119.

\item[] Stenstr\"om, A. (1994). {\em An Introduction to Spoken Interaction}, Longman, London.

\item[] Traum, D. J. Bos, R. Cooper, S. Larsson, I. Lewin, C. Matheson,
and M. Poesio (1999). `A model of dialogue moves and information state
revision'. {\sc Trindi} Project Deliverable D2.1, 1999.

\end{description}

\end{document}